\documentclass{article}

     \usepackage[final]{neurips_2020}

\usepackage[utf8]{inputenc} 
\usepackage[T1]{fontenc}    
\usepackage{hyperref}       
\usepackage{url}            
\usepackage{booktabs}       
\usepackage{amsfonts}       
\usepackage{nicefrac}       
\usepackage{microtype}      
\usepackage{graphicx}
\usepackage{subfigure}
\usepackage{listings}
\newcommand{\linestack}[1]{\def\arraystretch{1.0}\begin{tabular}[c]{@{}c@{}} #1 \end{tabular}}

\title{Neural Network Libraries: \\ A Deep Learning Framework Designed from Engineers' Perspectives}

\author{
 Takuya Narihira\thanks{Authors are listed in alphabetical order except for the first author.} \And
 Javier Alonsogarcia\And 
 Fabien Cardinaux\And
 Akio Hayakawa \And
 Masato Ishii \And
 Kazunori Iwaki\And 
 Thomas Kemp\And
 Yoshiyuki Kobayashi \AND
 Lukas Mauch\And
 Akira Nakamura \And
 Yukio Obuchi \AND
 Andrew Shin \And
 Kenji Suzuki\And
 Stephen Tiedmann\And
 Stefan Uhlich\And
 Takuya Yashima \And
 Kazuki Yoshiyama \AND
 {\large Sony Group Corporation}
}

\begin{document}

\maketitle
\begin{abstract}
  While there exist a plethora of deep learning tools and frameworks, the fast-growing complexity of the field brings new demands and challenges, such as more flexible network design, speedy computation on distributed setting, and compatibility between different tools. In this paper, we introduce Neural Network Libraries\footnote{https://nnabla.org}, a deep learning framework designed from engineer's perspective, with emphasis on usability and compatibility as its core design principles. We elaborate on each of our design principles and its merits, and validate our attempts via experiments.
\end{abstract}

\section{Introduction}
\label{sec_intro}

Deep learning has revolutionized the field of artificial intelligence, with state-of-the-art performances in image recognition (\cite{resnet, resnext}), speech recognition (\cite{speech}), and machine translation (\cite{mt}), just to name a few. Its application is not restricted to research area, and has taken up a substantial part of the real world platforms, such as automated driving and mobile applications. Moreover, with recent surge of generative models (\cite{gan, pggan, stylegan}), its potential as a tool for contents creation is also getting attention.

Deep learning research and development has seemingly formed a healthy ecosystem in which any researcher in the world can make a significant impact and contribution to the community, with rich amount of resources freely available (\cite{tensorflow2015-whitepaper, caffe, chainer, cntk, paszke2019pytorch}). On the other hand, as the field advances, more complexities and variations arise, and the demand for a more flexible and efficient tool grows stronger. For example, users need to define complex networks more concisely, and it is also necessary to easily handle static and dynamic computational graphs. Also, with the advent of massively large models (\cite{devlin-etal-2019-bert, NEURIPS2020_1457c0d6}), and the costs for accessing remote GPU servers skyrocketing, the ability to perform computation in a speedy manner, particularly on distributed setting, has become a pivotal factor.

Another issue that emerges from the massive expansion of deep learning tools is compatibility. With countless number of tools developed and released anew on a daily basis, it is possible that we end up with disjoint clusters of research and developments. Such can be concerning for a number of reasons; first, it can decelerate the speed of progress. Just as lack of appropriate translation between different languages can hinder communication, the absence of a protocol for bridging the gap between tools and frameworks can slow down the spreading of novel ideas. Second, in a similar manner, some of the innovations may fail to receive the deserved attention, due to their failure to merge with the mainstream cluster of tools. A tool to easily make it compatible with other tools and frameworks will alleviate such risks. 

We have developed and open-sourced Neural Network Libraries (NNL), focusing on the issues described above, namely usability and compatibility, particularly from the perspective of engineers. Our primary aim is to develop a deep learning framework that 1) enhances usability by flexible network design and speedy computation, and 2) that provides a wide range of compatibility, easily portable to and from other frameworks, thus lessening the unhealthy prospects described earlier. While such aims are equally critical for researchers as well, we attempt to approach the issues under the principle of ``Engineers First,'' as there already exists a plethora of research-oriented tools, with strikingly less amount of emphasis on engineering side. In this paper, we describe our design principles and their merits, how they are reflected in our framework, and further demonstrate them via experiments. We also briefly introduce Neural Network Console\footnote{https://dl.sony.com/}, a unique GUI-based development platform, which further promotes our design principles.

\section{Usability}
\label{sec_usability}
Design principle of Neural Network Libraries can best be described as "Engineers First". In other words, Neural Network Libraries is goal-oriented so that users can focus on constructing the network just by stacking the various layers sequentially.
Modern-day neural network frameworks utilize 2 types of computation graphs. One is a so-called 'static' graph, and the good example is TensorFlow (\cite{tensorflow2015-whitepaper}). Another is a 'dynamic' graph, adopted in PyTorch (\cite{paszke2019pytorch}). Neural Network Libraries supports both computation graphs, offering flexible usability to users. While we put forward the principle "Engineer First" here, researchers should also find it useful for their work since we offer reference implementations of many state-of-the-art models for various tasks such as image recognition or generative models.

\subsection{Building Neural Networks with Fewer Lines}
\label{sec_fewer_lines}
When users build a machine learning system, especially using deep learning, they need to 1) build a neural network, 2) prepare the data, and 3) train the network itself. Training takes a large amount of time, and so does data preparation. Also, training neural networks is likely to require several trial-and-errors. It is thus highly desirable to be time-efficient when building neural networks. 
In Neural Network Libraries, constructing the neural networks is easy and straightforward. 

\begin{lstlisting}[caption={Forward/Backward of the affine function},label={listing_affine},captionpos=b,language=python]
import numpy as np
import nnabla as nn
import nnabla.parametric_functions as PF

# Define input variable and computational graph
x = nn.Variable((16, 10), need_grad=True)
y = PF.affine(x, 5)

# Compute output for some random input
x.d = np.random.random(x.shape)
y.forward()

# Compute gradient with respect to input and parameters
y.backward()

# show all the trainable parameters assigned to the existing layers
nn.get_parameters()
\end{lstlisting}

As shown in Listing~\ref{listing_affine}, users simply need to prepare a data array and stack the layers to which the data is fed. Once the network is defined, it is represented as a computation graph internally, and calling \textbf{forward} method on the output data array executes computation. After forward computation is done, calling \textbf{backward} method performs backpropagation, where the gradients are calculated and subsequently stored to the corresponding arrays.

Neural Network Libraries offers three basic building blocks, namely, \textbf{Variables}, \textbf{Functions}, and \textbf{Parametric functions}. \textbf{Variables} represent data and their gradients with multi-dimensional arrays, \textit{e.g.}, tensors. \textbf{Functions} are mathematical operations that can be applied to variables. \textbf{Parametric functions} are functions accompanied with additional trainable parameters. Listing~\ref{listing_affine} shows example codes of using variables with a parametric function. Using these components, convolutional neural networks can also be easily implemented. A typical example is LeNet (\cite{lenet}), whose code sample is shown in Listing~\ref{python_lenet}, in Sec~\ref{sec_python_like_cpp_api}.

As can be seen in these examples, writing code with trainable layers (parametric functions) is straightforward since it does not require pre-defined layers and can be executed in a linear manner. In other words, users do not have to spend time on preparing the trainable parameters and assigning them to corresponding layers. All the trainable parameters are registered to a globally accessible dictionary. The last line in the Listing~\ref{listing_affine} shows the trainable parameters assigned to the existing layers.
Thus, users can construct the neural network in fewer lines, which enables them to quickly proceed to the training phase, and easily re-arrange the network design when necessary.

\subsection{Flexible Computation Methods}
\label{sec_static_dynamic}

\begin{figure}[h]
  \centering
  \includegraphics[width=0.80\linewidth]{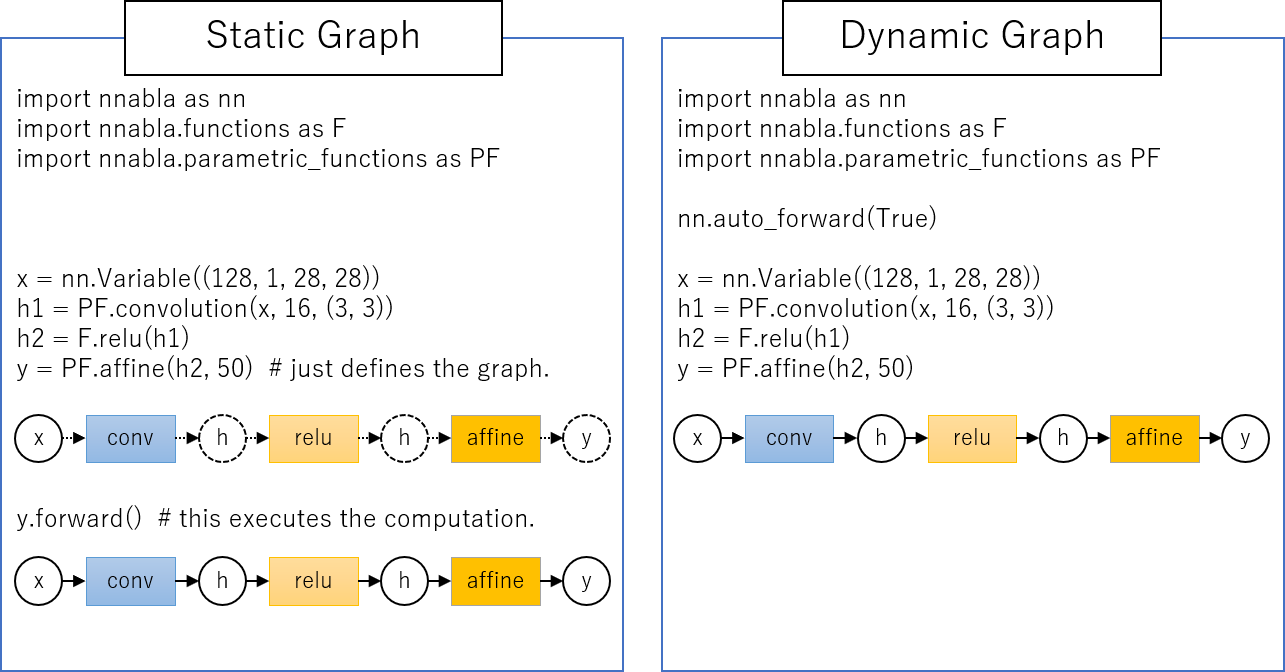}
  \caption{Illustration of static and dynamic graphs.}
  \label{static_and_dynamic_graph}
\end{figure}

Early neural network frameworks, such as Theano, Torch and TensorFlow utilize static computation graph. In static graph, users first need to define the entire graph and then use that graph for computation for each input data. The left block of Figure~\ref{static_and_dynamic_graph} shows a typical usage of static graph and its behavior. Actual computation is not executed at the time of graph definition, and the users need to call forward method explicitly. In static graph, the network architecture is completely fixed once defined. Instead, the computation speed is expected to be fast. 
Recent frameworks such as Chainer and PyTorch utilize dynamic computation graph in which users can dynamically change the network architecture and execute the computation on-the-fly. The right block of Figure~\ref{static_and_dynamic_graph} depicts its usage. Not only does it help users to understand what is happening inside the network, but it also provides more flexible usage of the neural network. For example, networks containing randomly dropping layers for each minibatch can be implemented. Switching to dynamic mode is also very simple, since it only requires addition of a single line as shown in Figure~\ref{static_and_dynamic_graph}. Necessity for modification of code is minimized.

\subsection{Speed Optimized with CuDNN / Distributed Training}
Most of the recent neural network frameworks utilize GPU to accelerate hardware for efficient training, and so does Neural Network Libraries. However, unlike other frameworks in which users need to assign tensors to GPU explicitly, Neural Network Libraries minimize users' efforts necessary to assign tensors to an appropriate device. Users simply need to switch the backend at the beginning of the code, which can be done by a single modification in extension context setting. (See Listing~\ref{backend_to_cudnn}.) Once device configuration is set, no other additional setup is needed and all \textbf{Variables} are automatically assigned to the chosen device. 

\begin{lstlisting}[caption={Switching backend in NNL},label={backend_to_cudnn},captionpos=b,language=python]
from nnabla.ext_utils import get_extension_context
nn.set_default_context(get_extension_context('cudnn'))
\end{lstlisting}

In deep learning, it is generally true that the more complex the model, the longer the training takes. This makes it particularly important to support data-parallel distributed training for deep learning frameworks. Aided by NCCL and MPI, Neural Network Libraries provides enhanced usability on distributed training setting as well, where inter-device communication and parallel computations can be easily implemented with addition of only a few lines as shown in Listing~\ref{distributed_training}.

\begin{lstlisting}[caption={Illustration of required modification for distributed training},label={distributed_training},captionpos=b,language=python]
import nnabla.communicators as C
ctx = get_extension_context("cudnn")
comm = C.MultiProcessDataParalellCommunicator(ctx)
comm.init()
...
params = [x.grad for x in nn.get_parameters().values()]
loss.backward(clear_buffer=True)
comm.all_reduce(params)
\end{lstlisting}

More details about the performance in term of computation speed can be found in Sec~\ref{sec_exp}.

\section{Compatibility}
\label{sec_comp}

\begin{figure}[h]
  \centering
  \includegraphics[width=0.99\linewidth]{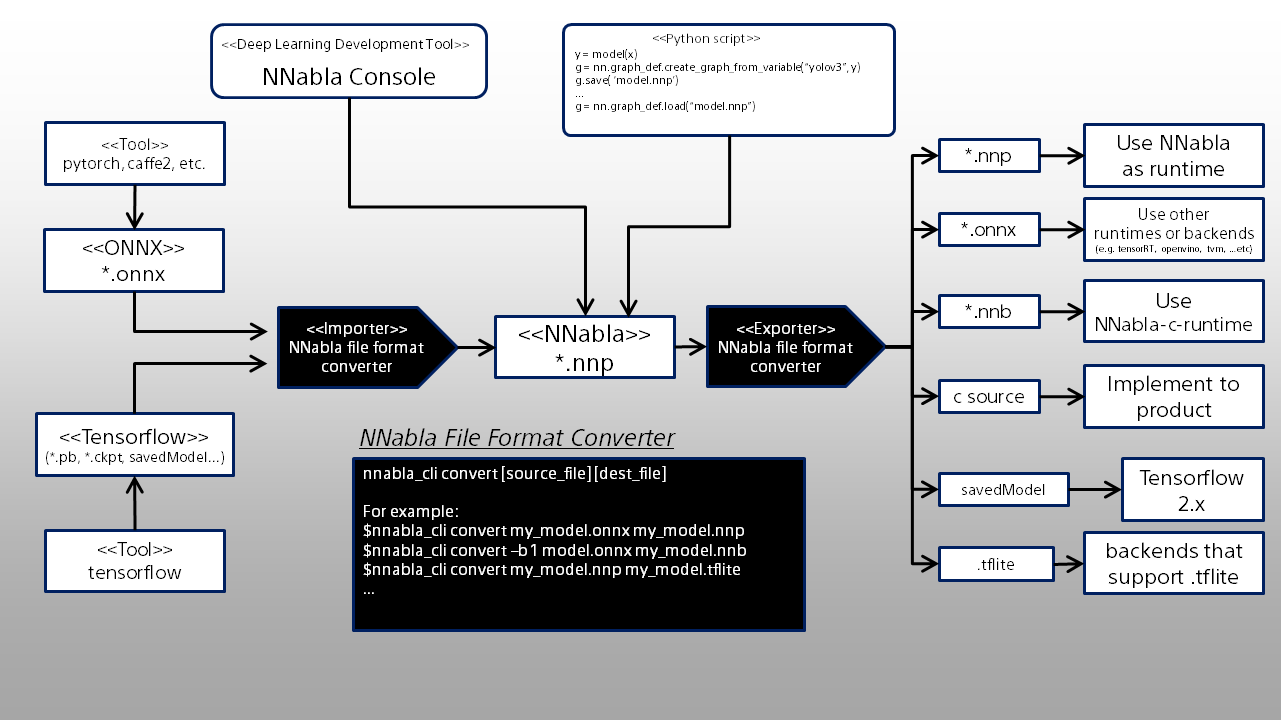}
  \caption{Overview of Neural Network Libraries' compatibility.}
  \label{portable}
\end{figure}

We provide tools for a variety of file format conversions, by which networks trained with Neural Network Libraries can be ported to other libraries including TensorFlow and Caffe, and vice versa, allowing for high development flexibility (Figure~\ref{portable}). Training a model on Neural Network Libraries generates an .\textit{nnp} file containing settings and parameters, which is portable to C++. In addition, we have also made it portable for ONNX\footnote{See \url{https://nnabla.readthedocs.io/en/latest/support_status.html} for specific function-level and model support status}. Trained networks can also be imported in our GUI module Neural Network Console, which we introduce later in this paper. With high portability, it is easy to deploy the trained models to applications of interest, or various neural network hardware-accelerated backends.

This file format converter uses protobuf defined in Neural Network Libraries as intermediate format. If ONNX file contains a function unsupported by Neural Network Libraries, it may cause error in conversion, so users may use querying commands provided by Neural Network Libraries to check whether it contains unsupported function. This converter also provides some intermediate process functionalities. We currently provide the following file format conversions:

\begin{itemize}
\itemsep0em 
\item NNP variations to valid NNP
\item ONNX to NNP and vice versa
\item NNP to NNB(Binary format for NNabla C Runtime)
\item NNP to Tensorflow frozen graph
\item Tensorflow checkpoint or frozen graph to NNP
\item NNP to TensorFlow Lite models
\item Experimental: Convert NNP to C Source code for NNabla C Runtime
\end{itemize}

Also, OpenCL (\cite{opencl}) extension can be easily used by simply building \textit{nnabla} and \textit{nnabla\_ext\_opencl} from source with OpenCL SDK setup. About 50 functions and solvers are currently implemented in OpenCL (FP-32 only).

\subsection{NNP Format}

NNP format contains the following information about the model preserved:
\begin{itemize}
\itemsep0em 
\item \textbf{NNablaProtoBuf}: Root message of NNabla network structure. This message can store GlobalConfig, TrainingConfig, Network(s), Parameter(s), Dataset(s), Optimizer(s), Monitor(s) and Executor(s).
\item \textbf{Variable}: Internal data structure to store tensor for Neural network I/O and parameters.
\item \textbf{GlobalConfig}: Configuration of environment for training or inference.
\item \textbf{TrainingConfig}: Configuration of training.
\item \textbf{Network}: Network structure.
\item \textbf{Parameter}: Special variable to store train result. (e.g Weight or Bias of affine layer). From the performance point of view, parameters can be saved in HDF5 (H5) format.
\item \textbf{Dataset}: Specify dataset for training.
\item \textbf{Optimizer}: Define network, dataset, and input/output variables for train.
\item \textbf{Monitor}: Define network, dataset, and input/output variables for monitor training status.
\item \textbf{Executor}: Define network and input/output variables for train.
\end{itemize}

\subsection{Python-like C++ API}
\label{sec_python_like_cpp_api}
It is easy to overlook the fact that the demand for flexible deployment of deep learning techniques is not limited to software applications, and its unfortunate consequence is the highly limited pool of resources designed from the low-level application or hardware perspective. Python-like C++ API is one of our attempts to alleviate such issue. While the capacity to program in C++ is a critical requirement in many applications integrating hardware, the complexity arising from the difference between C++ and Python programming has been an obstacle for a large number of developers. Our Python-like C++ API reflects the aspirations from such developers to be able to write in C++ as simply as in Python. 

Listing~\ref{python_lenet} and Listing~\ref{cpp_lenet} show how to implement LeNet (\cite{lenet}) in Neural Network Libraries with Python and Python-like C++ API respectively. It must be noted that our Python-like C++ API is able to implement the network with the same number of lines as in Python, with the difference in syntax confined to minimal extent. While many other frameworks including Neural Network Libraries provide low-level C++ API as well, it takes far more number of lines with much higher complexity to implement a network that can be written with a few lines in Python. Such verbosity and complexity frequently become the source of bug, as well as making the maintenance more troublesome. Our Python-like C++ API can perform both training and inference in C++ with the Python-like simplicity and syntax, and as such, it is expected to simplify and accelerate the various dimensions of C++-based applications development, encompassing programming, debugging, and maintenance.

\begin{lstlisting}[caption={Implementation of LeNet with Python},label={python_lenet},captionpos=b,language=python]
h = PF.convolution(x, 16, (5, 5), name="conv1")
h = F.max_pooling(h, (2, 2))
h = F.relu(h, inplace=False)
h = PF.convolution(h, 16, (5, 5), name="conv2")
h = F.max_pooling(h, (2, 2))
h = F.relu(h, inplace=False)
h = PF.affine(h, 50, name="affine3")
h = F.relu(h, inplace=False)
h = PF.affine(h, 10, name="affine4")
\end{lstlisting}

\begin{lstlisting}[caption={Implementation of LeNet with Python-like C++ API},label={cpp_lenet},captionpos=b,language=c++]
auto h = pf::convolution(x, 1, 16, {5, 5}, parameters["conv1"]);
h = f::max_pooling(h, {2, 2}, {2, 2}, true, {0, 0});
h = f::relu(h, false);
h = pf::convolution(h, 1, 16, {5, 5}, parameters["conv2"]);
h = f::max_pooling(h, {2, 2}, {2, 2}, true, {0, 0});
h = f::relu(h, false);
h = pf::affine(h, 1, 50, parameters["affine3"]);
h = f::relu(h, false);
h = pf::affine(h, 1, 10, parameters["affine4"]);
\end{lstlisting}

\subsection{Mixed Precision Training}
Mixed precision training (\cite{mixed}) is a method to train deep neural networks 
with half-precision floating points, which nearly halves the memory usage without sacrificing the performance. On top of memory usage, half-precision floating points also lead to the benefits of efficient bandwidth and significant speedup, although there is a potential side effect of unstable gradient computation due to quantization error. Some of the recent processors, such as Tensor Cores in NVIDIA Volta GPUs, support half-precision float computation, and NNL has also been designed to be compatible with mixed precision training. 

In NNL, mixed precision training can be used by setting \textit{type\_config} as \textit{half} in extension context setting. When using mixed precision training with NVIDIA Volta, storage (weights, activations, gradients) is performed in FP-16. Forward and back-propagation employ TensorCore, where batch normalization is in FP-32. Update is also performed in FP-32, although the weights are managed in both FP-16 and 32. Note that gradients may be too small to be represented in FP-16 and may necessitate loss scaling, which maintains a master copy of weights in FP-32. NNL provides automatic loss scaling updater class for such cases. Listing~\ref{nnabla_mixed_precision} shows example of mixed precision training in Neural Network Libraries.

\begin{lstlisting}[caption={Mixed precision training in Neural Network Libraries},label={nnabla_mixed_precision},captionpos=b,language=python]
# Use loss scaling to prevent underflow
loss_scale = 8
loss.backward(loss_scale)
solver.scale_grad(1. / loss_scale)  # some gradient clipping
solver.update()

# Use dynamic loss scaling to prevent overflow/underflow
scaling_factor = 2
counter = 0
interval = 2000
...
loss.backward(loss_scale, ...)
...
if solver.check_inf_or_nan_grad():
    loss_scale /= scaling_factor
    counter = 0
else:
    solver.scale_grad(1. / loss_scale) # some gradient clipping
    solver.update()
    if counter > interval:
        loss_scale *= scaling_factor
        counter = 0
    counter += 1
\end{lstlisting}

As we will see in Sec~\ref{sec_exp}, mixed precision training enables faster training than floating points, regardless of whether loss scaling is enabled or disabled.


\begin{figure}
\centering
\subfigure{\label{volta_train}\includegraphics[width=.55\linewidth]{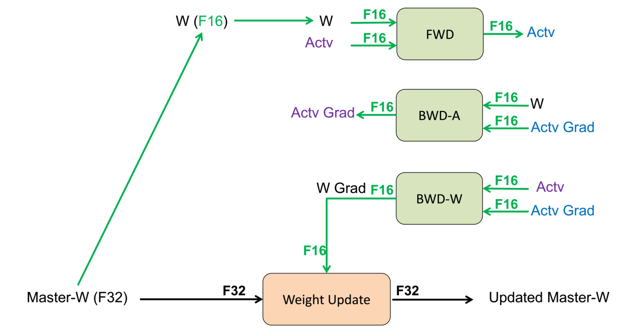}}
\subfigure{\label{volta_res}\includegraphics[width=.44\linewidth]{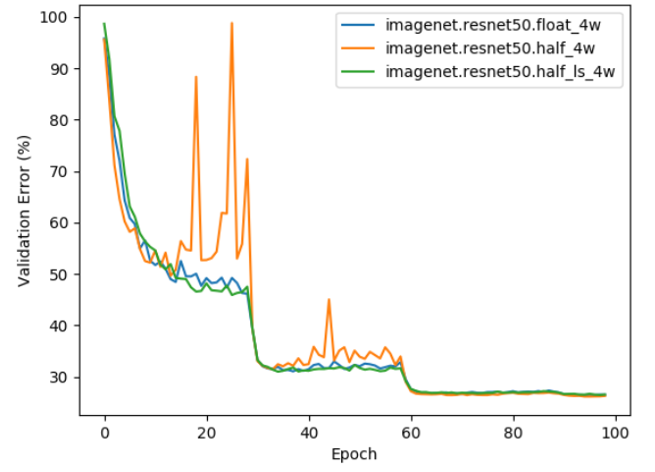}}
\caption{Volta training method (left) and the results of training resnet-50 with 4 volta-distributed training (right).}
\end{figure}

\section{Experiments}
\label{sec_exp}

To benchmark the performance of Neural Network Libraries, we conducted experiments with ImageNet dataset (ILSVRC 2012 classification dataset) \citep{imagenet_cvpr09}, which is widely used to evaluate the performance of deep neural networks. We implemented various kinds of popular DNN architectures and evaluated their training speed and test accuracy. For fair comparison with other existing libraries, we follow the setup adopted in NVIDIA Deep Learning Examples\footnote{https://github.com/NVIDIA/DeepLearningExamples}. We used DGX-1 with 4 GPUs (NVIDIA V100) for the training and measured training time for 90 and 250 epochs. Since NVIDIA Deep Learning Examples do not report the training time with 4 GPUs, we doubled the reported training time with 8 GPUs and compared our results with them. Our implementation used for this experiment is publicly available in our GitHub repository\footnote{https://github.com/sony/nnabla-examples/tree/master/imagenet-classification}. Note that this implementation completely relies on NVIDIA's GPUs, and uses NVIDIA's data processing library, called DALI, which works only on Linux.

We first compared the training time of Neural Network Libraries with PyTorch and TensorFlow, the two most popular libraries in deep learning. We trained ResNet-50, one of the most popular network architectures, using Neural Network Libraries and compared its training time with that reported in NVIDIA Deep Learning Examples. Table \ref{tab:exp_speed} shows the training time as well as how much it is reduced by adopting the mixed precision. Neural Network Libraries achieves competitive speed when compared to other libraries due to efficient distributed training over multiple GPUs as well as utilization of mixed precision. Figure~\ref{volta_train} illustrates how we trained the model, along with the results of training.

We also evaluated the training time and validation error rate with various model architectures. Table \ref{tab:exp_various} shows the results with ResNet variants (ResNeXt \citep{resnext} and SE-ResNet/SE-ResNeXt \citep{seresnext}), and Table \ref{tab:exp_lightweight} shows those with several popular lightweight models, such as MobileNet-V3 \citep{howard2019searching} and EfficientNet \citep{pmlr-v97-tan19a}.

\begin{table}[t]
    \centering
    \begin{tabular}{|c|ccc|}
    \hline
         & FP-32 & Mixed precision & Speedup by mixed precision  \\
    \hline
    PyTorch &  24 h & 10 h & x2.3 \\
    TensorFlow & 20 h & 7 h & x3.0 \\
    NNabla & 23.3 h & 7.4 h & x3.1 \\
    \hline
    \end{tabular}
    \caption{Training time of ResNet-50 for 90 epochs with ImageNet dataset.}
    \label{tab:exp_speed}
\end{table}

\begin{table}[h]
    \centering
    \begin{tabular}{|c|ccc|}
    \hline
    Network architecture & \linestack{Training time\\ (90 epochs)} & \linestack{Training time\\ (250 epochs)} & \linestack{Validation error\\ (250 epochs)} \\ 
    \hline
    ResNet-18 & 6.7 h & 16.1 h & 28.3 \% \\
    ResNet-50 & 7.44 h & 20.2 h & 21.6 \% \\
    ResNeXt-50 & 12.1 h & 33.8 h & 21.0 \% \\
    SE-ResNet-50 & 15.0 h & 42.2 h & 21.2 \% \\
    SE-ResNeXt-50 & 19.7 h & 55.7 h & 20.1 \% \\
    \hline
    \end{tabular}
    \caption{Training time and validation error for variations of ResNet architecture}
    \label{tab:exp_various}
\end{table}

\begin{table}[h]
    \centering
    \begin{tabular}{|c|cc|}
    \hline
    Network architecture & \linestack{Training time\\ (350 epochs)} & \linestack{Validation error\\ (350 epochs)} \\
    \hline
    MobileNet-V3 small & 5.5 h & 32.9 \% \\
    MobileNet-V3 large & 7.6 h & 24.9 \% \\
    EfficientNet-B0 & 50.0 h & 23.7 \% \\
    EfficientNet-B1 & 79.5 h & 21.9 \% \\
    EfficientNet-B2 & 95.5 h & 20.9 \% \\
    EfficientNet-B3 & 148.9 h & 19.4 \% \\
    \hline
    \end{tabular}
    \caption{Training time and validation error for lightweight models}
    \label{tab:exp_lightweight}
\end{table}

\section{Extensions}
\label{sec_ext}
While we designed our framework under the principle of ``Engineers First,'' this does not imply in any way that we neglect to account for the usage by people with lesser or no engineering background. On the contrary, taking full advantage of our engineer-oriented approach, we strove to widen the range of people who can benefit from deep learning technology, by providing novel ways to deploy them. The representative work of such endeavor is Neural Network Console, our GUI-based deep learning development platform, which is also directly compatible with Neural Network Libraries. We have also highly committed to incorporating deep learning into entertainment field, especially contents creation. As such, a wide array of generative models have been constantly made available with pre-trained weight parameters for immediate usage off-the-shelf.

\subsection{Neural Network Console}

Despite the current buzz around AI and deep learning, people with little or no background in programming have found it challenging to approach it, and frequently ended up on the surface level without understanding its essence by performing design and training neural networks themselves. Even for experienced programmers, developers, and researchers, it requires a very large number of evaluation cycles and can be highly time-consuming to prepare, pre-process, train, and evaluate. 

We provide GUI-based deep learning development tool, Neural Network Console (NNC), that enables users to develop their own neural networks more easily. With GUI, users can design neural network structure with visual programming using layers (function blocks) by simple drag and drop. Parameters are automatically calculated and errors can be confirmed immediately. As such, it is particularly useful for learning the concept and learning to design the networks. Also, since all trials are recorded automatically, it is easy to analyze the performance and revert to old records if necessary. Results of experiments are listed and can be compared to past trials. For a classification task, it displays a confusion matrix. Since it checks for validation error, number of parameters, and multiply-adds in real time, users can benefit from significant amount of speed up in evaluation cycle. Highly complex networks and experimental settings such as ResNet-152, generative adversarial networks (GANs) (\cite{gan}), or semi-supervised setting can be implemented as well.

NNC also supports automatic structure search function, which searches for optimal neural network structure automatically by repeating experiments with varying network structures. Multiple network structures are evaluated, simultaneously optimizing for accuracy and computational complexity. Users can select from multiple optimization results. Thus, automatic structure search enables a significant efficiency in optimization of neural networks and is useful for development of embedded applications.

On top of supporting novice developers, NNC provides plugin features that enable users to develop more flexible layers, data processing, and analytical methods. Since user can use Python to design their own plugin, it is easy to develop complex deep learning models such as explainable AI (XAI). For example, we provide a variety of XAI-related plugins, including Grad-CAM (\cite{grad-cam}), LIME (\cite{lime}), and SGD influence (\cite{NEURIPS2019_5f146156}). Users can examine these cutting-edge techniques simply on GUI.

\begin{figure}
\centering
\subfigure[Interface for designing networks]{\label{fig:a}\includegraphics[width=.85\linewidth]{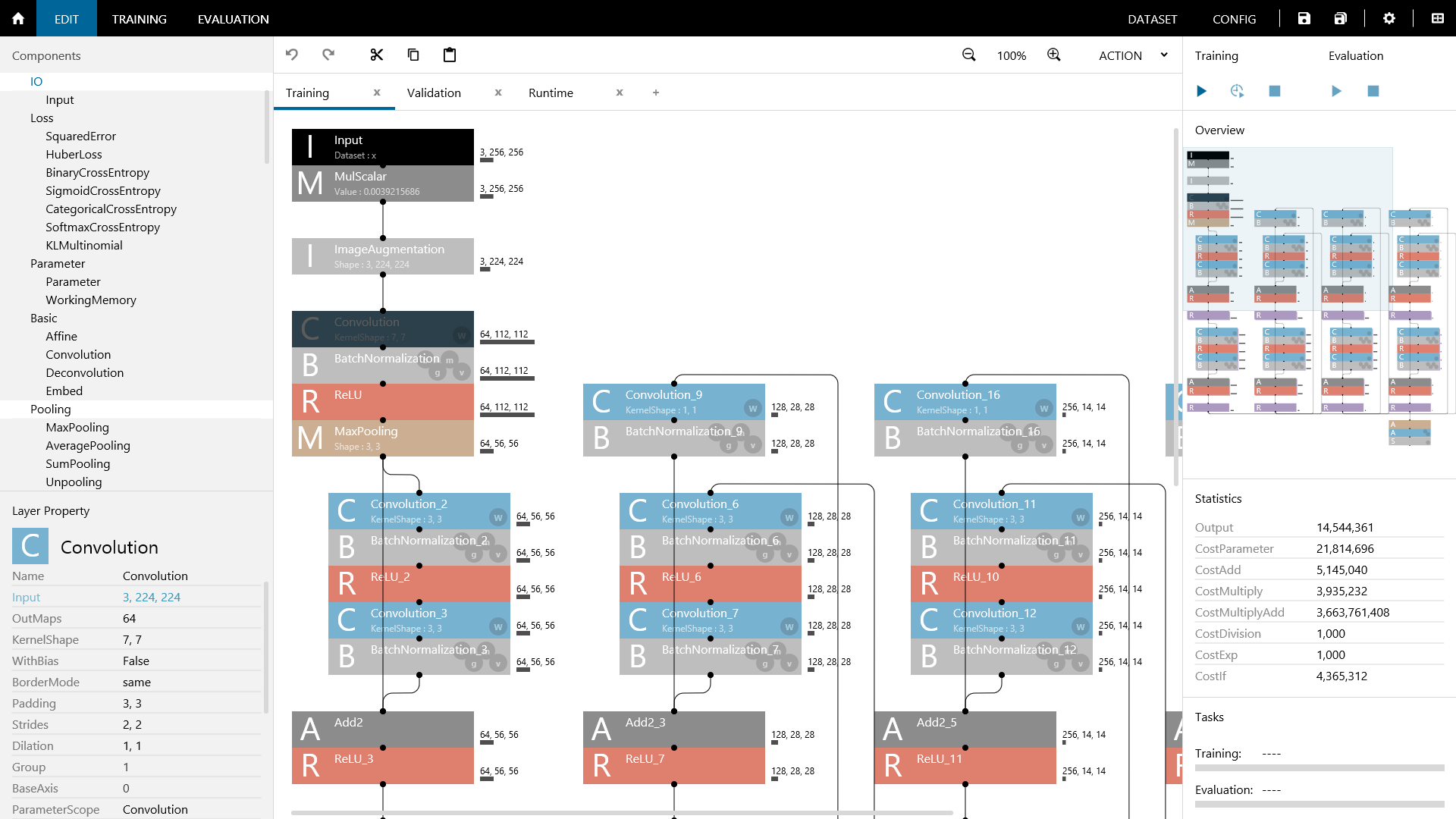}}
\subfigure[Interface for training and evaluation]{\label{fig:b}\includegraphics[width=.85\linewidth]{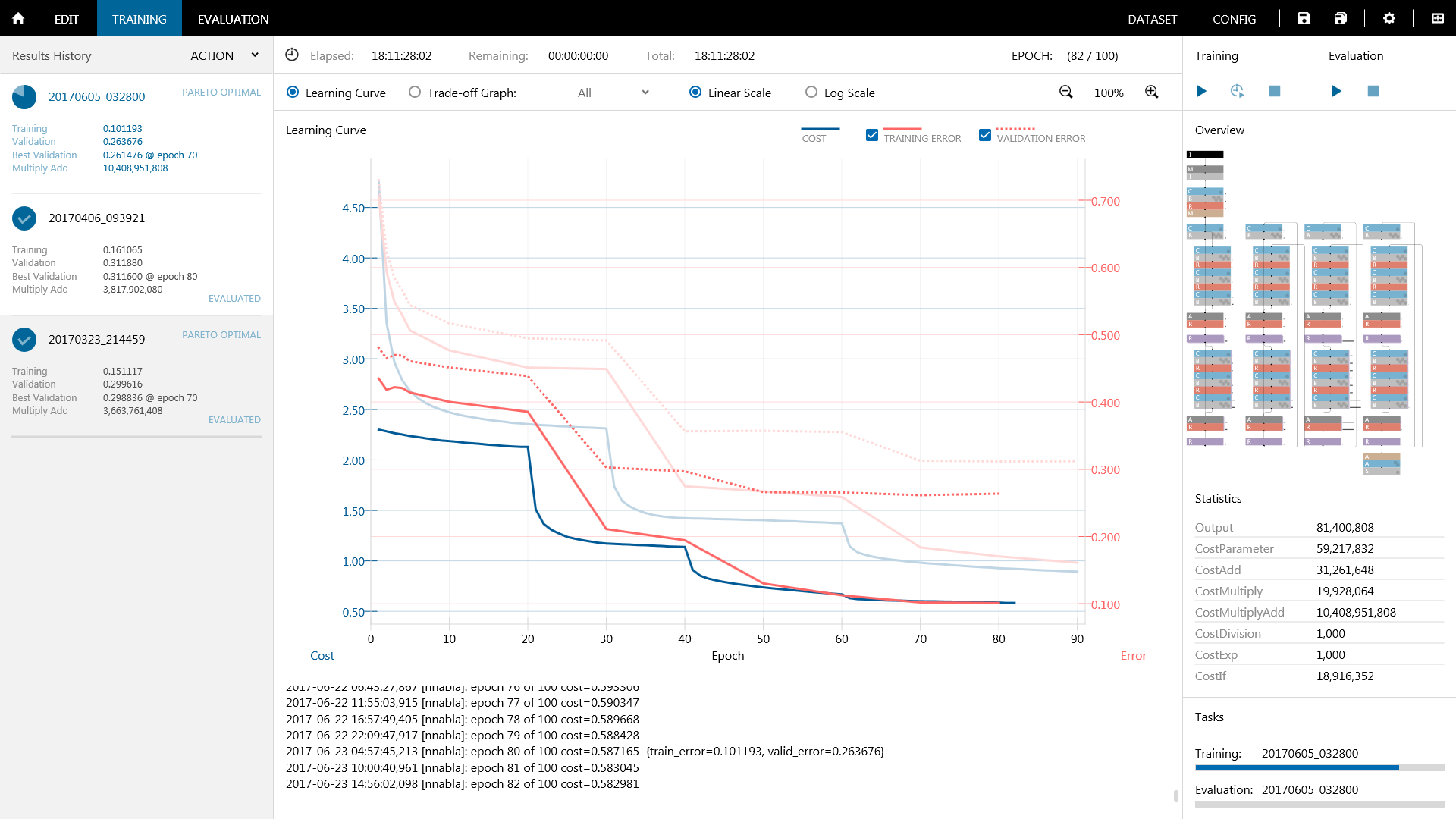}}
\caption{User interface of Neural Network Console.}
\label{console_ui}
\end{figure}

NNC is mutually portable with NNL. Thus, if users want to visually confirm whether the network designed in NNL is correct, they can simply import the exported file from NNL (.\textit{nntxt} format) into NNC. It can also be useful when they want to footprint the computational workload of the networks designed in NNL. Figure~\ref{console_ui} shows example screenshots of Neural Network Console's GUI for training and inference phases respectively.

\label{sec_console}

\subsection{Contents Creation}
Over the past years, generative models have evolved dramatically from merely generating low-resolution images (\cite{gan}), to generating high-resolution images that are indistinguishable from actual objects (\cite{pggan, stylegan, stylegan2}). Such dramatic evolution opens up an unbounded possibility for a variety of creative tasks. We have noted that such immense potential of generative models will be clearly beneficial particularly for contents creators, and have striven to provide an environment, in which our users can benefit directly from most important and up-to-date models, readily available with pre-trained weights. Our current lineup\footnote{https://github.com/sony/nnabla-examples/tree/master/GANs} encompasses a wide range of generative models, from milestone models such as DCGAN (\cite{dcgan}), CycleGAN (\cite{cyclegan}), pix2pix (\cite{pix2pix}), progressive growing of GANs (\cite{pggan}), StyleGAN2 (\cite{stylegan2}), to more recent task-specific models, such as ESR-GAN (\cite{esrgan}), TecoGAN (\cite{tecogan}), StarGAN (\cite{stargan}), InstaGAN (\cite{instagan}), ReenactGAN (\cite{reenactgan}) etc. We plan to aggressively continue to add new models to our lineup.

While generative models can equip contents creators with powerful tools, its benefits and accessibility are still limited to users with moderate level of technical expertise and computational resources. In order to overcome this obstacle and widen the range of users who can benefit from these models, we have prepared a lineup of interactive demos\footnote{https://github.com/sony/nnabla-examples\#interactive-demos}, where users can easily experience each model, without having to worry about the technical details, using freely available GPU resources from Google Colab\footnote{https://colab.research.google.com/}. We also plan to actively add new models to this lineup, and hope to engage more users in our attempts to broaden the user base of deep learning technology.

\section{Conclusion}
We have developed a deep learning framework, Neural Network Libraries, with engineer-oriented design principles, putting strong emphasis on usability and compatibility. The features of Neural Network Libraries reflect such design principles, and we demonstrate via experiments that they are efficient. Finally, we have also developed extensions and applications to spread the benefits of Neural Network Libraries to non-engineers as well.

\subsubsection*{Acknowledgments}

We would like to express our genuine gratitude for the contributions made by the members of Sony R\&D Center Stuttgart Laboratory, China Software Center of Sony China, Software Architecture Division (SARD) of Sony India Software Center, and many other voluntary contributors. We would also like to thank our users who have provided constructive feedback and insights.

\small
\bibliographystyle{plainnat}
\bibliography{neurips_2020}

\end{document}